\pdfoutput=1

\documentclass[11pt]{article}

\usepackage[final]{acl}

\usepackage{times}
\usepackage{latexsym}

\usepackage[T1]{fontenc}

\usepackage[utf8]{inputenc}

\usepackage{microtype}

\usepackage{inconsolata}

\usepackage{multirow}
\usepackage{array}
\usepackage{booktabs}
\usepackage{graphicx}
\usepackage{amsmath}

\usepackage{amsmath,amsfonts,bm}

\def\eqref#1{equation~\ref{#1}}

\def\1{\bm{1}}

\DeclareMathAlphabet{\mathsfit}{\encodingdefault}{\sfdefault}{m}{sl}
\SetMathAlphabet{\mathsfit}{bold}{\encodingdefault}{\sfdefault}{bx}{n}

\usepackage{makecell}

\newcommand{\EMT}[1]{\textcolor{blue}{{\small [#1 --EMT]}}}

\newcommand{\Dicio}{\textsc{Dicio}}
\newcommand{\Larousse}{\textsc{Larousse}}
\newcommand{\Oxford}{\textsc{Oxford}}
\newcommand{\DRAE}{\textsc{DRAE}}
\newcommand{\Duden}{\textsc{Duden}}

\newcommand{\mBARTCC}{m\textsc{BART-cc25}}
\newcommand{\mBART}{m\textsc{BART-50}}
\newcommand{\mTT}{m\textsc{T5}}

\newcommand{\LLama}{\textsc{LLama-2}}
\newcommand{\Mistral}{\textsc{Mistral}}
\newcommand{\LLamathree}{\textsc{Llama-3}}

\newcolumntype{C}[1]{>{\centering\let\newline\\\arraybackslash\hspace{0pt}}m{#1}}
\newcolumntype{L}[1]{>{\raggedright\let\newline\\\arraybackslash\hspace{0pt}}m{#1}}

\title{Multilingual Definition Modeling}

\author{
    Edison Marrese-Taylor\textsuperscript{1,2},
    Erica K. Shimomoto\textsuperscript{1},
    Alfredo Solano\textsuperscript{2},
    Enrique Reid\textsuperscript{2}\\
    National Institute of Advanced Industrial Science and Technology\textsuperscript{1}\\
    The University of Tokyo\textsuperscript{2}\\
    \tt edison.marrese@aist.go.jp, kidoshimomoto.e@aist.go.jp\\
    \tt asolano@weblab.t.u-tokyo.ac.jp, erarvin2007@gmail.com
}

\begin{document}
\maketitle

\begin{abstract}
In this paper, we propose the first multilingual study on definition modeling. We use monolingual dictionary data for four new languages (Spanish, French, Portuguese, and German) and perform an in-depth empirical study to test the performance of pre-trained multilingual language models on definition modeling of monosemic words when finetuned on this data. Furthermore, we use a zero-shot approach to test the multilingual capabilities of two popular chat-based Large Language Models (LLMs) in the task. Results show that multilingual language models can perform on-pair with English but cannot leverage potential cross-lingual synergies, with LLMs generally offering better performance overall. A comprehensive human evaluation of the LLM-generated definition highlights the zero and few-shot capabilities of these models in this new task, also showing their shortcomings. Finally, we show that performance on our task via BERTScore strongly correlates to the performance on multilingual LLM benchmarks, suggesting that our task offers a viable compute-constrained, stable and natural alternative to these.

\end{abstract}

\section{Introduction}

Definitions play a key role in effective communication and language understanding. However, creating high-quality and precise definitions requires time and effort, given the complexity of languages and how they are constantly evolving. 

Aiming to suffice this necessity, the definition modeling task was first proposed by \citet{norasetDefinitionModelingLearning2017}. The goal is to estimate the probability of a textual definition given a word being defined. This task can be applied to generate definitions for known and new words, assisting language learners, as well as serving as a tool for language preservation and description~\cite{kabiri2020evaluating, dimas-furtado-etal-2024-dore}.

Since its conception, several approaches have tackled this task, showing that it gives an arguably more transparent view of how much a model captures syntax and semantics. 

So far, existing approaches for this task are trained on a corpus of word-definition pairs, to be later tested on how well they generate definitions for words not seen during training. Current approaches \cite{norasetDefinitionModelingLearning2017,gadetsky-etal-2018-conditional,niLearningExplainNonStandard2017a,ishiwatari-etal-2019-learning} are mainly encoder-decoder based, in which a contextual representation for a word/phrase, encoded using a variety of features, is used to generate the definition.
Despite the progress, previous work has predominantly focused on definition modeling in English. Concretely, we find just a handful of papers that work on definition modeling for languages other than English, namely  \citet{reidVCDMLeveragingVariational2020} for French, \citet{dimas-furtado-etal-2024-dore} for Portuguese and \citet{kongCrossLingualDefinitionGeneration2020} for Chinese. We also find that dictionary data for French and Italian have been released in the past few years, but that there are no approaches that leverage them so far \cite{hathoutGLAFFLargeVersatile2014,hathoutWiktionnaireWikicodeGLAWIfied2016,sajousENGLAWIHumanMachineReadable2020,calderoneHybridMethodStress2017}. Finally, while \citet{kabiri2020evaluating} has covered nine different languages in their study, they did not explore multilingual capabilities of models, focusing on performance of monolingual models on the polysemy aspect of words.
Moreover, we note that these datasets are mostly based on crowdsourced dictionaries, where quality can be a concern.

In light of these issues, in this paper, we present the first \textit{truly} multilingual study on definition modeling, introducing datasets obtained from not-crowdsourced dictionaries in French, Spanish, Portuguese, and German, with a total of approximately 460K new terms entries and 730K definitions, thus substantially extending language coverage for the task. We combine our collected data with existing corpora to train models and lay out the first set of experiments on multilingual definition modeling. As our main focus is to analyze the potential synergy between different languages, we focus our experiments on monosemic words, hoping to explore polysemy in future work.

Furthermore, we also test a selection of Large Language Models (LLMs) on our collected data, probing their multilingual abilities in this task through zero-shot and few-shot approaches \cite{brownLanguageModelsAre2020,kojimaLargeLanguageModels2022}. We evaluate performance based on three different automatic metrics, i.e., BLEU, BERTScore and COMET, and through human evaluation. 

Our results show that current multilingual language models such as mBART \cite{liuMultilingualDenoisingPretraining2020,tangMultilingualTranslationExtensible2020} and mT5 \cite{xueMT5MassivelyMultilingual2021} can perform on-pair with English when fine-tuned on single-language dictionaries. They can also deliver consistent performance when trained in multiple languages, but are unable to leverage cross-lingual synergies to consistently improve performance. As for LLMs, we show that prompting techniques, either in the zero-shot or few-shot scenarios, offer generally better performance than multilingual language model fine-tuning, even in the case of LLMs. However, we also observe that the output language can be difficult to control and that techniques such as in-context learning do not lead to significantly better quality output. 

Finally, we show empirical evidence suggesting that per-language performance in our task measured via BERTScore is strongly correlated with performance on multilingual benchmarks. Opposite to these, which are often translations of datasets in English, our task can offer a compute-constrained, static and natural alternative to assess the proficiency of LLMs on a given language. We release our code and data on GitHub\footnote{\url{https://github.com/epochx/defmod}}.

\section{Related Work}

Our paper is primarily related to the seminal work by \citet{norasetDefinitionModelingLearning2017,hill-dictionary-2016}, in which a model is tasked with generating a definition for a word given its respective embedding or with mapping dictionary definitions to lexical representations of words, respectively. Later work has proposed improvements and extensions, introducing techniques and datasets to address shortcomings. For example, \citet{gadetsky-etal-2018-conditional} address polysemy and presents a dataset from Oxford Dictionaries, where each definition is supplemented with context sentences. \citet{niLearningExplainNonStandard2017a} proposed an approach for automatically explaining slang English terms in a sentence and introduced another dataset extracted from The Urban Dictionary. \citet{ishiwatari-etal-2019-learning} proposed to further rely on local and global contexts to help the model disambiguate and generate better definitions. Different from our approach, their work focused on unfamiliar words and phrases (many of which are polysemous words), rarely used idioms, or emerging entities, relying on Wikipedia as a source.


More recently, \citet{huangDefinitionModellingAppropriate2021} proposed to study the problem of definition specificity, creating a method for tuning a model to account for hyper-focused (over-specific) or highly general (under-specific) definitions. Finally, \citet{chenUnifiedModelReverse2022} proposed to unify the seminal ideas of reverse dictionary and definition modeling in a single model, with the goal of helping better understand word sense and embeddings.
\citet{peritiAutomaticallyGeneratedDefinitions2024} proposed to directly finetune an LLM specifically for definition modeling. 

Previous work discussed so far has mostly focused on definitions in the English language. There are a few exceptions, such as the work of \citet{reidVCDMLeveragingVariational2020}, who presented the first study on definition modeling for the French language with the release of a dataset collected from Le Petit Robert, and the work of \citet{dimas-furtado-etal-2024-dore}, whose main goal was to collect a dataset for Portuguese definition generation, leveraging several models. Their approach is, however, still monolingual.

We also note that over the past few years, dictionary datasets in several languages derived from Wiktionary have been released, including English \cite{sajousENGLAWIHumanMachineReadable2020}, French \cite{hathoutGLAFFLargeVersatile2014,hathoutWiktionnaireWikicodeGLAWIfied2016}, Italian \cite{calderoneHybridMethodStress2017}. \citet{ylonenWiktextractWiktionaryMachineReadable2022} collected data for all available Wiktionaries. These datasets, were not accompanied by models that leverage them, and we were unable to find other work using them to train or evaluate models on definition modeling. We think one critical point here is that Wiktionaries are built on the base of crowdsourcing, where quality could be a concern. 

Finally, we find previous work aiming to generate definitions in specialized fields, e.g., scientific and medical terms \cite{augustGeneratingScientificDefinitions2022}, biomedical terminology definition \cite{liuGraphineDatasetGraphaware2021,huangUnderstandingJargonCombining2022}, or financial terminology \cite{jhiradEvaluatingLargeLanguage2023}.

\section{Data}
\label{sec:data}

We are interested in assessing the capabilities of multilingual models to describe words in multiple languages. Therefore, we aim to collect high-quality data for ``traditional'' dictionary definitions, where most common words in a given language are contained. This approach deviates from recent previous work in definition modeling, where the focus has been on domains of ``uncommon'' words like slang terms \cite{niLearningExplainNonStandard2017a} or rarely used idioms and emerging entities \cite{ishiwatari-etal-2019-learning} and specific domains such as scientific or financial terms \cite{augustGeneratingScientificDefinitions2022,liuGraphineDatasetGraphaware2021,huangDefinitionModellingAppropriate2021, jhiradEvaluatingLargeLanguage2023}, and only in English.

To support our study, we utilize data for three of the Indo-European languages with the largest number of speakers, according to Wikipedia: Spanish (485 M native speakers), Portuguese (236 M native speakers), and French (80 M native speakers). As we are particularly interested in studying potential performance gains when training on multiple languages, we focus on these in particular, as they all derive from the Romance branch and thus share many similarities. Finally, we also consider German, another Indo-European language but from the Germanic family with 75 M native speakers. We deem German as the ideal candidate to experiment together with existing datasets for the English language.

\begin{table}[h!]
    \centering
    \scalebox{0.9}{
    \begin{tabular}{l@{\hspace{0.15cm}} r@{\hspace{0.15cm}} c@{\hspace{0.15cm}} c@{\hspace{0.15cm}}}
    \toprule
        \bf Dataset (language) & \bf Terms   & \bf Defs. & \bf Mono.\\
        \midrule
        \Oxford{} - en          & 36,767  & 122,319     & 44.07\% \\
        Le Petit Robert - fr    & 33,507  & -           & -        \\
        \midrule
        \DRAE{} - es (ours)     & 82,386  & 159,863   & 60.85\%  \\
        \Dicio{} - pt (ours)    & 191,499 & 279,985   & 77.41\%  \\
        \Larousse{} - fr (ours) & 72,486  & 130,408   & 70.06\%  \\
        \Duden{} - de (ours)    & 110,444 & 164,086   & 73,17\%  \\
        \bottomrule
    \end{tabular}
    }
    \caption{Summary of our collected datasets, compared to prior relevant corpora. In the table, Mono. stands for Monosemic, i.e. terms with a single definition. We compare against \Oxford{} and Le Petit Robert (data from \citet{reidVCDMLeveragingVariational2020}), which we regard as existing datasets extracted from ``traditional'' dictionaries.}
    \label{table:data}
\end{table}

For each language, we start by collecting word lists, which are put together from various online sources, including but not limited to MUSE \cite{lample2018word}. 

We found inflections of verbs (e.g., tenses) and adjectives (e.g., gender) are present and relied on spacy\footnote{\url{https://spacy.io/}} to identify word lemmas, utilizing the \textit{en\_core\_web\_sm} model for English as well as the \textit{core\_news\_sm} model for German, Spanish, Portuguese and French. After obtaining the lemmatized version of each entry in our data, we kept the example where the original term matches the lemmatized word, similarly to \citet{ishiwatari-etal-2019-learning}.

In order to obtain good quality definitions for the terms collected, we work with traditional, well-known and readily available dictionaries. Concretely, we rely on the Dictionary of the Spanish language (``Diccionario de la lengua española - DLE'') developed and maintained by the Royal Spanish Academy (Real Academia Española - RAE) for Spanish, Dicio for Portuguese, the Larousse Dictionary for French, and the Duden Dictionary for German. Collecting data from online dictionaries can involve accessing copyrighted information. In this sense, our approach follows previous work \cite{gadetsky-etal-2018-conditional,niLearningExplainNonStandard2017a,ishiwatari-etal-2019-learning,reidVCDMLeveragingVariational2020,dimas-furtado-etal-2024-dore} in making sure that the data can indeed be downloaded/distributed. Based on our analysis of the Terms of Service for the consulted dictionaries, we will not directly distribute our datasets, but instead release recipes on how to reconstruct them based on our compiled word lists. Please see \S \ref{sec:appendix_data} for further details about the elaboration of our datasets.

Table \ref{table:data} summarizes the main characteristics of the data we collected, compared against existing resources in English and French, where we can see that our collected data is substantially richer, both in terms of the number of terms and definitions.

\section{Experimental Setup}
\label{sec:experimental_setup}

\paragraph{Data} An important distinction between our corpus and recent prior work is that our collected data does not contain examples of word usage. While we acknowledge the importance of such context to disambiguate a specific meaning for a given term in the case of polysemy, we opt to explore \textit{monosemic} words as a first step into the study of multilinguality for definition modeling. We thus select the subset of terms that exhibit only a single meaning, which in practice we achieve by simply selecting the terms with a single definition entry. The final word lists are split into the 80/10/10 ratio. Table \ref{table:splits} below shows the exact details of our splits.

In addition to our data, we also use the previously released dataset built from Oxford Dictionaries \cite{gadetsky-etal-2018-conditional} (\Oxford). Though we were also interested in using the French data collected by \citet{reidVCDMLeveragingVariational2020}, we found it was not available at the time of writing this paper.

\begin{table}[h]
    \centering
    \scalebox{0.9}{
    \begin{tabular}{lrrr}
        \toprule
        \textbf{Dataset}    & \bf Train  & \bf Valid & \bf Test  \\
        \midrule
        \Oxford{} (en)         & 15,770  & 6,884  & 6,834  \\
        \midrule
        \DRAE{} (es)           & 32,834  & 4,104  & 4,105  \\
        \Dicio{} (pt)          & 118,591 & 14,824 & 14,824 \\
        \Larousse{} (fr)       & 31,224  & 3,903  & 3,904  \\
        \Duden{} (de)          & 55,521  & 6,940  & 6,941      \\  
        \bottomrule
        \end{tabular}
        }
        \caption{Details of the size of each split for our collected data filtered for polysemy, compared against \Oxford.}
        \label{table:splits}
        \vspace{-0.5cm}
    \end{table}

\paragraph{Model Finetuning}  We train models based on mT5 \cite{xueMT5MassivelyMultilingual2021}. This model is a multilingual version of T5 \cite{raffelExploringLimitsTransfer2020}, a generative Transformer pre-trained on a Common Crawl-based dataset covering 101 languages and including a data for a variety of tasks which have been converted into a text-to-text format. We begin by training monolingual baselines for each language, finetuning each dataset. These models give us a set of robust baselines. We then train combining data for several languages. We utilize the ``large'' model (m\textsc{T5}) (approx. 1.5 B parameters) with the prefix ``\textit{Define in $\{$language$\}$}'', where the variable \textit{$\{$language$\}$} indicates the language in which we would like the model to generate (English, Spanish, Portuguese, French, German). During training, we randomly sample (word, definition) pairs from the data involved and use these examples to create multilingual batches. In preliminary experiments, we also tested alternative multilingual models, but found that mT5 performed better overall.

\paragraph{Model Prompting} Given their success in solving several NLP tasks, we also evaluate the ability of instruction-tuned or instruction-tuned LLMs to provide definitions for words. We consider white-box models including Llama 2 \cite{touvron2023llama}, Llama 3.1 \cite{grattafiori2024llama3herdmodels}, and Mistral v1 models \cite{jiang2023mistral}. Specifically, we utilize \textit{Llama-2-13b-chat-hf} (\LLama), \textit{Llama-3.1-8B-Instruct} (\LLamathree{}), \textit{Mistral-7B-Instruct-v0.1} (\Mistral), respectively. Via preliminary experiments using 4-bit quantization with QLoRA \cite{dettmers2023qlora} on a subset of languages and models, we observed that the effects of quantization were bounded $<$1\% of the full model performance, and affected the datasets evenly, and thus adopted this strategy for all our LLM experiments. 

We test two settings: (1) zero-shot, where the model is directly asked to generate the definition of the word, and (2) few-shot, where we incorporate term-definition examples in the prompt before requesting the definition for the target term. These shots are randomly sampled from the training data and kept constant across examples. 

Prompting LLMs can lead to substantially different outputs depending on how instructions are crafted. For our experiments, we kept the input to the model as simple as possible, always using the prompt \textit{``Define the $\{$language$\}$ word `$\{$term$\}$'. Use only $\{$language$\}$ to reply.''}, where $\{$language$\}$ and $\{$term$\}$ are variables denoting the target language and the term to define. We repeat each experiment 3 times with different random seeds. In the case of few-shot, we also sample a different set of (term, definition) tuples from the train portion of the dataset to use as context.

\paragraph{Evaluation} Previous work in definition modeling has mainly utilized n-gram overall metrics such as BLEU \cite{papineni-etal-2002-bleu} and METEOR \cite{banerjee-lavie-2005-meteor}. As the latter is language-specific, here we report BLEU, relying on the sacrebleu\footnote{\url{https://github.com/mjpost/sacrebleu}} implementation \cite{post2018call}. Previous work \cite{reidVCDMLeveragingVariational2020,huangDefinitionModellingAppropriate2021} has also shown that metrics based on n-gram overlap cannot capture nuance in the definitions generated by models. As an alternative, approaches have relied on metrics based on Machine Learning, such as BERTScore \cite{zhang*BERTScoreEvaluatingText2019}. Following \citet{jhiradEvaluatingLargeLanguage2023} we also experiment with COMET \cite{reiCOMETNeuralFramework2020}, using the \textit{wmt22-comet-da} model, which has multilingual support as it is built on top of XLM-R \cite{conneauUnsupervisedCrosslingualRepresentation2020}. 

To evaluate the ability of the LLMs to generate definitions in the target language, we rely on a fasttext-based language classification model \cite{grave-etal-2018-learning}. Here, compare the desired target language with the highest probability language as predicted by this model.  We report the percentage of examples where the prediction matches the desired target language (Compliance).

Finally, we also perform human evaluation on a subset of the definitions generated by our models. We split our study into two subsets: Romance and Germanic languages, with particular emphasis on the former, due to their linguistic similarities and relative higher number of first language speakers. We recruit three volunteer native speakers of Brazilian Portuguese, European French, and Spanish, and one native speaker of American English, and German. To make sure the evaluations are consistent across languages, we designed an evaluation scheme based on a Likert scale with 5 values (1 to 5), which we crafted based on a detailed examination of the outputs of the models. 

Please see \S \ref{sec:appendix_experimental_setup} for more details on models, setup, training and inference, including the results of our study on quantization and perfomance the model used for language identification. For details about our evaluation scheme, please refer to \S \ref{sec:human_evaluation_details}.

\section{Results}

\subsection{Automatic Evaluation}

Table \ref{table:results_multilingual} summarizes the results of our multilingual experiments using \mTT. We observe that this model can perform consistently when trained in multiple languages, as measured by COMET and BERTScore. When it comes to BLEU scores, although we observe little to no improvements over the monolingual baselines in the validation set, we see a more consistent behavior on the test portion. However, we note that the definitions generated by our models are often correct but expressed using different surface forms compared to the gold standard, which can lead to low BLEU scores.

Overall, our results show that even for languages from the same family, where we often find that similar words are associated to similar or related meanings (e.g. words sharing the same Latin root in the case of the Romance languages), exposing models such multilingual lexical data did not help them generalize across languages. However, training models in this fashion does not lead to degraded per-language performance either. A possible explanation for these observations is that a significant portion of the similar term-definition pairs across languages lie in polysemic terms.

\begin{table}[t]
    \footnotesize
    \begin{tabular}{cl c@{\hspace{0.25cm}} c@{\hspace{0.25cm}} c@{\hspace{0.25cm}}}
        \toprule
        \bf Dataset & \bf Training Data & \bf BL     & \bf CMT   & \bf BS \\
        \midrule
        \multirow{2}{*}{\shortstack{\Duden\\(de)}}
            & de                & \bf 0.939  & \bf 0.346 & \bf 0.664  \\
            & de + en           & 0.507      & 0.345     & 0.656  \\
        \midrule
        \multirow{4}{*}{\shortstack{\Oxford\\(en)}}
            & en                & \bf 1.330  & \bf 0.345 & 0.813  \\
            & en + fr           & 0.038      & \bf 0.345 & 0.778  \\
            & en + de           & 0.863      & 0.344     & \bf 0.839  \\
            & en + fr + es + pt & 0.532      & \bf 0.345 & \bf 0.839  \\
        \midrule
        \multirow{5}{*}{\shortstack{\DRAE\\(es)}}
            & es                & 7.700      & \bf 0.384 & 0.743  \\
            & es + fr           & 8.084      & 0.382     & 0.743  \\
            & es + pt           & 9.158      & 0.382     & 0.746  \\
            & es + pt + fr      & \bf 9.244  & 0.381     & \bf 0.747  \\
            & en + fr + es + pt & 5.503      & \bf 0.384 & 0.727  \\
        \midrule
        \multirow{6}{*}{\shortstack{\Larousse\\(fr)}}
            & fr                & 0.490      & \bf 0.300 & 0.685  \\
            & fr + es           & 0.936      & 0.293     & 0.686  \\
            & fr + pt           & 0.587      & 0.298     & 0.675  \\
            & fr + en           & 0.104      & 0.294     & 0.644  \\
            & fr + es + pt      & \bf 2.660  & 0.292     & \bf 0.705  \\
            & fr + en + es + pt & 0.748      & 0.293     & 0.682  \\
        \midrule
        \multirow{5}{*}{\shortstack{\Dicio\\(pt)}} 
            & pt                & 11.539     & 0.352     & 0.732  \\
            & pt + es           & 18.626     & 0.352     & \bf 0.748  \\
            & pt + fr           & 4.969      & \bf 0.353 & 0.682  \\
            & pt + fr + es      & \bf 18.856 & 0.352     & \bf 0.748  \\
            & en + fr + es + pt & 4.542      & 0.351     & 0.685  \\ 
        \bottomrule 
    \end{tabular}
    \caption{Results of our multilingual experiments based on \mTT{}, where BL is short for BLEU, CMT is short for COMET, and BS is short for BERTScore.}
    \label{table:results_multilingual}
    \vspace{-0.5cm}
\end{table}

\begin{table}[ht!]
    \footnotesize
    \begin{tabular}{c@{\hspace{0.15cm}} l@{\hspace{0.25cm}} c@{\hspace{0.2cm}} c@{\hspace{0.25cm}} c@{\hspace{0.25cm}} c@{\hspace{0.25cm}} c@{\hspace{0.25cm}}}
        \toprule
        \textbf{Dataset} & \textbf{Model} & \bf S & \bf BL  &  \bf BS  &\bf CMT & \bf Cmp. \\
        \midrule
        \multirow{6}{*}{\shortstack{\Duden\\(de)}} 
            & \multirow{2}{*}{\LLama}
                & 0 & 0.180 & 0.612 & \bf \underline{0.439} & \bf0.685 \\
                & & 5 & \underline{\bf0.758} & \underline{\bf0.637} & 0.378 & 0.604 \\
            \cmidrule{2-7}
            & \multirow{2}{*}{\Mistral}
                & 0 & 0.159 & 0.621 & \underline{\bf0.411} & \bf0.762 \\
                & & 5 & \bf0.233 & \bf0.623 & 0.362 & 0.628 \\
            \cmidrule{2-7}
            & \multirow{2}{*}{\LLamathree}
                & 0 & 3.261 & 0.676 & 0.580 & \bf 0.998 \\
                & & 5 & \underline{\bf7.967} & 0.695 & 0.558 & 0.982 \\
        \midrule
        \multirow{6}{*}{\shortstack{\Oxford\\(en)}}
            & \multirow{2}{*}{\LLama}
                & 0 & 0.949 & 0.826 & \underline{\bf0.553} & - \\
                & & 5 & \underline{\bf3.906} & \underline{\bf0.851} & 0.537 & - \\
            \cmidrule{2-7}
            & \multirow{2}{*}{\Mistral}
                & 0 & 1.183 & 0.837 & \underline{\bf0.557} & 0.999 \\
                & & 5 & \underline{\bf1.821} & \underline{\bf0.842} & 0.529 & - \\
            \cmidrule{2-7}
            & \multirow{2}{*}{\LLamathree}
                & 0 & 1.588	& 0.835 & 0.568 & - \\
                & & 5 & \underline{\bf3.571} & 0.847 & 0.537 & - \\
        \midrule
        \multirow{6}{*}{\shortstack{\DRAE\\(es)}}
            & \multirow{2}{*}{\LLama}
                & 0 & 0.271 & 0.617 & \underline{\bf0.504} & \underline{\bf0.996} \\
                & & 5 & \underline{\bf1.360} & \underline{\bf0.675} & 0.470 & 0.877 \\
            \cmidrule{2-7}
            & \multirow{2}{*}{\Mistral}
                & 0 & 0.324 & 0.650 & \underline{\bf0.458} & 0.703 \\
                & & 5 & \bf0.559 & \underline{\bf0.663} & 0.440 & \underline{\bf0.852} \\
            \cmidrule{2-7}
            & \multirow{2}{*}{\LLamathree}
                & 0 & 1.855 & 0.670 & 0.545 & \bf 0.999 \\
                & & 5 & \underline{\bf3.067} & 0.683 & 0.530 & 0.974 \\
       \midrule
       \multirow{6}{*}{\shortstack{\Larousse\\(fr)}}
            & \multirow{2}{*}{\LLama}
                & 0 & 0.326 & 0.631 & \bf0.473 &\bf 0.970 \\
                & & 5 & \underline{\bf1.350} & \underline{\bf0.676} & 0.469 & 0.966 \\
            \cmidrule{2-7}
            & \multirow{2}{*}{\Mistral}
                & 0 & 0.327 & 0.648 & \underline{\bf0.441} & 0.775 \\  
                & & 5 & \underline{\bf 0.626} & \underline{\bf0.664} & 0.425 & \bf0.951 \\
            \cmidrule{2-7}
            & \multirow{2}{*}{\LLamathree}
                & 0 & 3.443 & 0.676 & \bf0.543 & \bf 0.999 \\
                & & 5 & \underline{\bf6.635} & \bf0.685 & 0.513 & 0.950 \\
       \midrule
       \multirow{6}{*}{\shortstack{\Dicio\\(pt)}}
            & \multirow{2}{*}{\LLama}
                & 0 & 0.143 & 0.682 & \bf0.513 & \bf0.969 \\
                & & 5 & \bf1.828 & \bf0.683 & 0.462 & 0.883 \\
            \cmidrule{2-7}
            & \multirow{2}{*}{\Mistral}
                & 0 & 0.146 & 0.689 & 0.475 & 0.630 \\
                & & 5 & \underline{\bf4.030} & \underline{\bf0.709} & \underline{\bf0.501} & \underline{\bf0.926} \\
            \cmidrule{2-7}
            & \multirow{2}{*}{\LLamathree}
                & 0 & 0.457 & 0.683 & 0.542 & \bf 0.998 \\
                & & 5 & \underline{\bf2.298} &\bf0.715 & \bf0.549 & 0.982 \\
       \bottomrule
    \end{tabular}
    \caption{Results of our experiments with LLMs, where S indicates the number of shots, and Cmp. is short for compliance, the \% of cases where the answer is in the correct language. BL, BS and CMT are short for BLEU, BERTScore and COMET, respectively. Numbers in bold indicate best results for each language-model combination. Underlined results indicate differences against zero-shot are statistically significant at level $\alpha = 0.05$.}
    \label{table:zero_shot}
\end{table}

Table \ref{table:zero_shot} summarizes the results of our experiments with LLMs. Compared to our finetuning experiments, LLMs generally offer better performance in terms of COMET and BERTScore and are competitive in terms of BLEU. One important issue here is that LLMs are sometimes unable to follow instructions, which, in our case, often leads to the model generating outputs in English. 

When it comes to differences due to in-context learning (ICL), although our results may suggest that this leads to improvements, these were not always statistically significant across metrics and languages, showing that ICL may not be as effective for this task as suggested by previous work. 

\paragraph{Role of Prompt Language} As explained in \S \ref{sec:experimental_setup}, the prompt utilized in our experiments with LLMs is provided in English. Since this may bias models toward producing output in this language, we performed a study using a prompt in Spanish on the \DRAE{} dataset. In this case, we use the following prompt template: ``Define la palabra `$\{$term$\}$'. Utiliza solo español en tu respuesta.'', a loose translation of our English prompt, where `$\{$term$\}$' is a placeholder for the actual word. We follow the same methdology and test on \LLama{} and \LLamathree{}, on the 0-shot and 5-shot scenarios. Table \ref{table:results_prompt_language} summarizes our results and compares them against our baseline prompt in English. As shown in the table, results show that prompting models in English leads to consistent superior performance accross all metrics. This validates our choice to prompt models solely in English, and supports our observations during early experimentation. These results also suggest that more recent models offer increased abilities to follow instructions in languages other than English. However, a significant gap still exists. 

\begin{table}[t]
\footnotesize
    \begin{tabular}{lcccccc}
        \toprule
        \textbf{Model} & \textbf{S} & \textbf{L} & \textbf{BL} & \textbf{BS} & \textbf{CMT} & \textbf{Cmp.} \\
        \midrule
        \multirow{4}{*}{\LLama{}}
            & \multirow{2}{*}{0}
                     & en & \bf 0.271 & \bf 0.617 & \bf 0.504 & \bf 0.996 \\
                &    & es & 0.186 & 0.601 & 0.502 & 0.814 \\
        \cmidrule(lr){2-7}
            & \multirow{2}{*}{5}
                    & en & \bf 1.360 \bf & \bf 0.675 & \bf 0.470 & \bf 0.877 \\
                &   & es & 0.019 & 0.526 & 0.409 & 0.078 \\
        \midrule
        \multirow{4}{*}{\LLamathree{}}
            & \multirow{2}{*}{0}
                    & en & \bf 1.855 & \bf 0.670 & \bf 0.545 & - \\
                &   & es & 0.303 & 0.619 & 0.532 & - \\
        \cmidrule(lr){2-7}
            & \multirow{2}{*}{5}
                    & en & \bf 3.067 & 0.683 & \bf 0.530 & \bf 0.974 \\
                &   & es & 2.885 & \bf 0.711 & 0.529 & 0.908 \\
        \bottomrule
        \end{tabular}
    \caption{Performance on Spanish definition modelling when using prompts in English (en) and Spanish (es), tested on \DRAE{}, were S and L indicate the number of shots and prompt language respectively. BL, BS and CMT are short for BLEU, BERTScore and COMET, respectively. Numbers in bold indicate best results per model.}
    \label{table:results_prompt_language}
    \vspace{-0.3cm}
\end{table}

\paragraph{Dataset Contamination} A key aspect to consider when analizing our results is the the possibility that portions of our benchmark have appeared in the training data of the LLMs that we tested \cite{sainz-etal-2023-nlp,deng-etal-2024-unveiling,li-etal-2024-open-source}. An important point here is to first clarify what ``contamination'' should mean in the context of our task. We think exposing LLMs to actual word-definition pairs during training can and should be considered a kind of contamination. However, if models are simply exposed to examples of words/terms in context, this should not count as a case of contamination. Moreover, the degree of exposure of the LLMs to these contexts is highly relevant to our task and should be studied. One possibility is to work only with fully open-source LLMs, which would enable us to analyze the training data. This is left for future work.

\subsection{Human Evaluation}

We further evaluate the output of the 5-shot experiments on all of our languages for \LLama{} and \Mistral{} via a human study. We use 200 randomly-sampled terms, for one random seed across languages, totaling 400 definitions to be evaluated per language. We additionally require the annotators to provide us with an assessment of the quality of the gold standard definitions we collected via the same 1-5 Likert score defined in \S \ref{sec:experimental_setup}. During the whole evaluation procedure, we allow the annotators to see gold-standard definitions and consult external sources if needed. Table \ref{table:human_eval_1} summarizes our results.

\begin{table}[h]
    \centering
    \footnotesize
    \begin{tabular}{c@{\hspace{0.25cm}} l@{\hspace{0.25cm}} c@{\hspace{0.25cm}} c@{\hspace{0.25cm}} c@{\hspace{0.25cm}} c@{\hspace{0.25cm}}}
        \toprule
        \
        \multirow{2}{*}{\bf Dataset} & \multirow{2}{*}{\bf Model} & \multirow{2}{*}{\bf Score} & \multicolumn{3}{c}{\bf IAA} \\
        \cmidrule(lr){4-6}
         & & & $\kappa$  & $\tau$ & \bf EM  \\
        \midrule
        \multirow{3}{*}{\shortstack{\Duden\\(de)}}
            & \LLama      & 2.050 & -      & -      & -     \\
            & \Mistral    & 1.950 & -      & -      & -     \\
            & Data        & 3.825 & -      & -      & -     \\
        \midrule
        \multirow{3}{*}{\shortstack{\Oxford\\(en)}}
            & \LLama      & 4.830 & -      & -      & -     \\
            & \Mistral    & 4.710 & -      & -      & -     \\
            & Data        & 4.925 & -      & -      & -     \\
        \midrule
        \midrule
        \multirow{3}{*}{\shortstack{\DRAE\\(es)}}
            & \LLama      & 2.303 & 0.520 & 0.788 & 0.675 \\
            & \Mistral    & 1.815 & 0.616 & 0.836 & 0.785 \\
            & Data        & 4.565 & 0.134 & 0.148 & 0.600  \\
        \midrule
        \multirow{3}{*}{\shortstack{\Dicio\\(pt)}}
            & \LLama      & 1.812 & 0.337 & 0.594 & 0.633 \\
            & \Mistral    & 1.352 & 0.281 & 0.446 & 0.777 \\
            & Data        & 2.967 & 0.164 & 0.920 & 0.355 \\
        \midrule
        \multirow{3}{*}{\shortstack{\Larousse\\(fr)}}
            & \LLama      & 2.720 & 0.449 & 0.652 & 0.563 \\
            & \Mistral    & 2.133 & 0.581 & 0.731 & 0.700 \\
            & Data        & 4.972 & 0.083 & 0.125 & 0.965 \\
        \bottomrule
    \end{tabular}
    \caption{Results of our human evaluation on our best results with LLMs, in terms of average Likert scores, along with their IAA measured via average Cohen's Kappa ($\kappa$), Kendall's Tau ($\tau$) and exact match (\textbf{EM}).}
    \label{table:human_eval_1}
    \vspace{-0.5cm}
\end{table}

\paragraph{Inter-annotation agreement} We report Cohen's Kappa ($\kappa$) scores, which we regard as a lower bound in our case, as it assumes label independence. Since our Likert scores are inherently a ranking, we also report Kendall's Tau ($\tau$), which accounts for this and offers a more realistic agreement measure. Finally, we also report the percentage of examples where the annotators have an exact match (\textbf{EM}) in their answers. For languages where we have more than two evaluators, we report the average value of all posible pairwise combinations.

As shown in Table \ref{table:human_eval_1} we first see a clear gap in performance between English and in the rest of the languages, which is in agreement with our automatic evaluation. These results further confirm that all three models struggle with Portuguese and German, at least in comparison with the other European languages, despite reportedly being trained on a similar amount of text as is the case of \LLama{}, where the training mixture contains 0.17\% de, 0.16\% fr, 0.13\% es \cite{touvron2023llama}. 

Our human evaluation also shows substantial variation in terms of Likert scores for the quality of the gold standard definitions, suggesting that \Dicio{} and \Duden{} may offer comparatively lower quality data. In contrast, although results on \Larousse{} exhibit particularly low $\kappa$ and $\tau$ values, we see an \textbf{EM} of 0.97. Given the high average score of 4.9, this shows that although the annotators may disagree about the exact score for a given definition, overall they think that definitions are of excellent quality.

Finally, we allowed the annotators to provide us with comments, and one concrete issue that was brought to our attention was that these dictionaries would often offer just a synonym as a definition for a given term. This phenomenon may respond to historical reasons (originally, dictionaries were printed) or be due to the editorial of each company. Nevertheless, if we regard these scores as an upper bound for the human evaluation, we find that in all cases there is still a significant gap between this and the performance of the LLMs. We performed an additional human study to gain insight into the effect of ICL and the effectiveness of COMET as a metric, please see \S \ref{sec:human_evaluation_details} for details.

\subsection{Multilingual benchmark comparison}

We study the relationship between our proposed task and existing datasets used to measure the multilingual performance of LLMs. Concretely, we compare per-language performance for each one of our metrics against multilingual versions of HellaSwag \cite{zellers-etal-2019-hellaswag}, ARC-Challenge \cite{clark2018think} and MMLU \cite{hendrycks2021measuring}, provided by \citet{lai-etal-2023-okapi} via the \textit{lm-evaluation-harness}\footnote{\url{github.com/EleutherAI/lm-evaluation-harness}} platform \cite{eval-harness}. 
Together, these datasets contain multiple choice questions spanning a wide variety of tasks, including adversarial commonsense natural language inference, abstraction and reasoning ability acquisition via demonstrations, and questions derived from diverse fields of knowledge. 

We found that per-language performance in our task, measured via BERTScore, strongly correlates with all of our selected multilingual benchmarks,
with average values of 0.94, 0.95 and 0.87 for \LLama{}, \Mistral{} and \LLamathree{}, respectively. Please see Table \ref{table:correlation_details} for the details of all the correlation values we obtained, and \S \ref{sec:detail_correlation_benchmarks} for the exact benchmark performance we measured. Compared to these, which are machine-translations of the originals in English, we think the definition modeling task can offer a natural, stable alternative to assess the proficiency of LLMs in a given language. 

Furthermore, to understand the computational trade-offs of using our task instead of our studied multilingual benchmarks, we compare the correlation between performance on the latter and in our datasets, versus subsamples of different sizes (from 0.1\% to 5\%). For each size, we sample 10 times and report average correlation and confidence intervals via Gaussian-based asymptotic approximation.

\begin{figure}[h!]
    \centering
    \includegraphics[width=0.95\columnwidth]{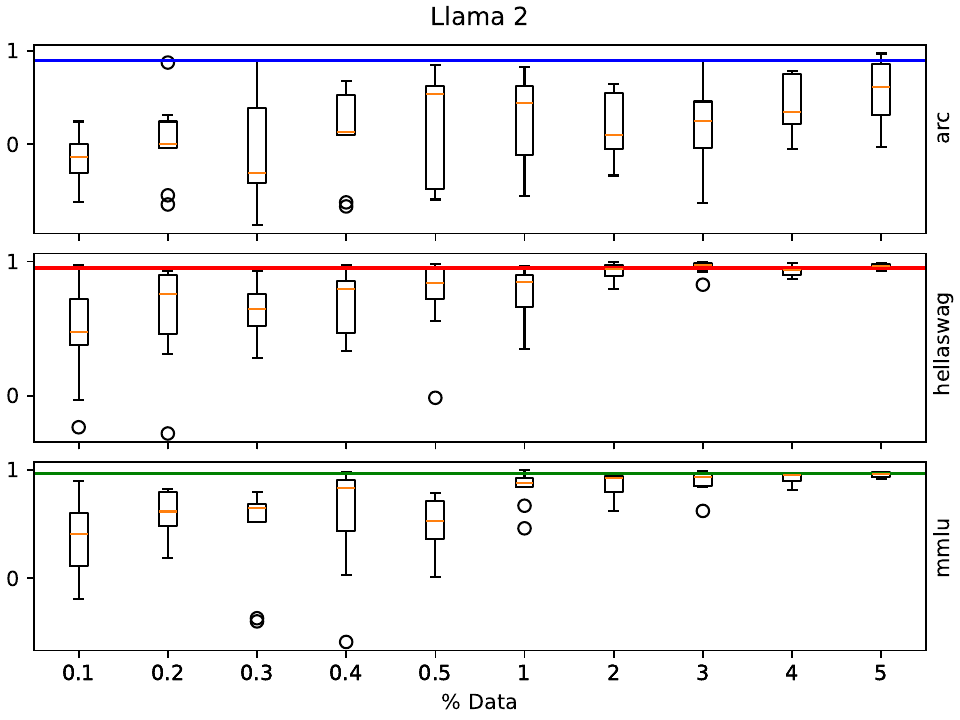}\\
    \includegraphics[width=0.95\columnwidth]{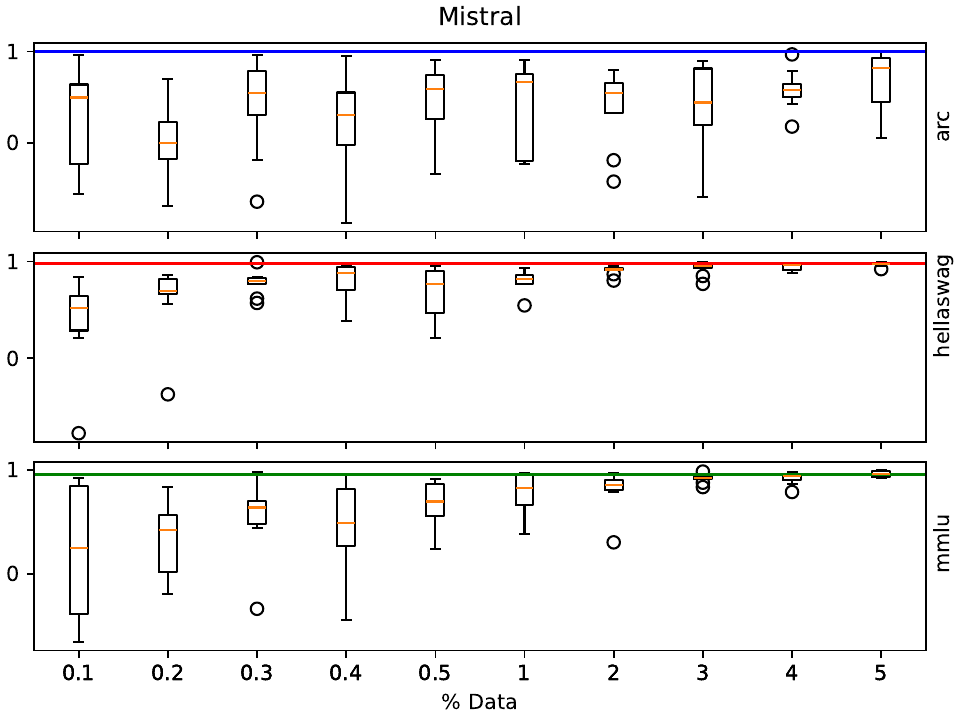}
    \includegraphics[width=0.95\columnwidth]{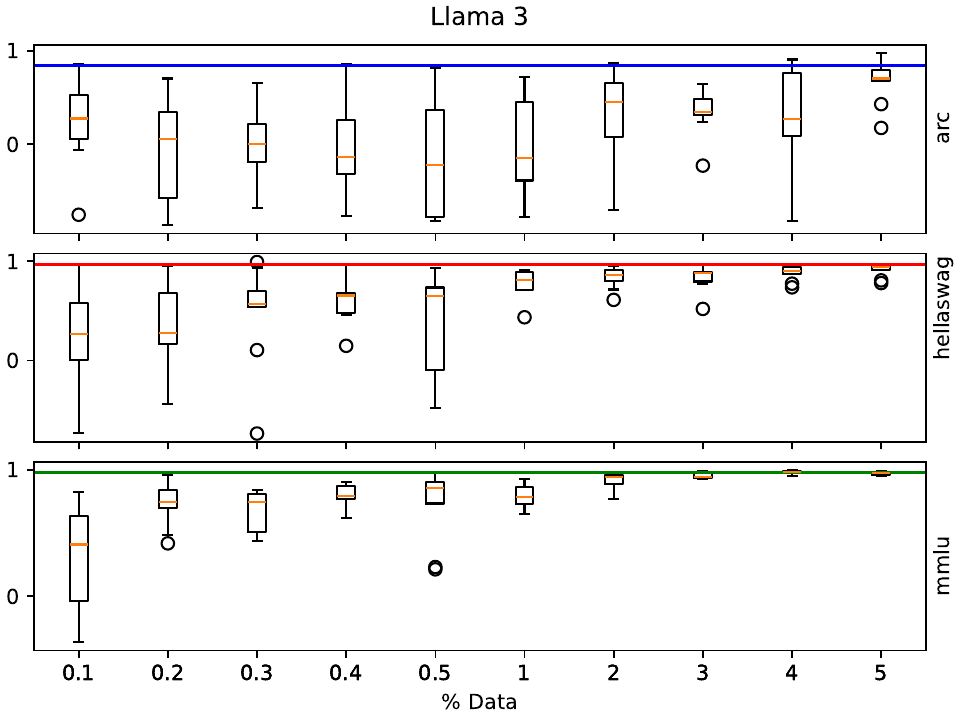}
    \caption{Pearson correlation between performance on multilingual benchmarks and on our task via BERTScore (solid lines), versus the correlation between the same benchmarks and random samples of different sizes for our three tested models \LLama{} (top), \Mistral{} (center), and \LLamathree{} (bottom).}
    \label{figure:benchnmark_correlation}
    \vspace{-0.5cm}
\end{figure}

Figure \ref{figure:benchnmark_correlation} summarizes our findings. The  computational budget required for our dataset is roughly equivalent to 0.5\% of MMLU, 18\% of ARC-Challenge or 36.7\% or HellaSwag in number of tokens (more details available in Table \ref{table:flops}). This means that definition modeling can offer an efficient alternative to measuring the LLM performance on a specific language, particularly when compared with computationally-intensive benchmarks like MMLU. Our datasets can also be combined with modern sampling approaches for bechmmark score approximation such as IRT \cite{polo2024tinybenchmarks}, leading to further efficiency gains.

\subsection{Comparison with prior work}

\begin{table}[t]
    \centering
    \scalebox{0.87}{
    \begin{tabular}{lccc}
        \toprule
        \bf Model & \bf BL & \bf CMT & \bf BS \\
        \midrule
        Full \Oxford{}&  &  &  \\
        \midrule
        LoG-CAD (\citeauthor{ishiwatari-etal-2019-learning}) & 25.19 & - & - \\
        VCDM (\citeauthor{reidVCDMLeveragingVariational2020}) & 27.38 & 0.570$^\diamondsuit$ & - \\
        T5 + RR (\citeauthor{huangDefinitionModellingAppropriate2021}) & 26.52 & - & - \\
        LLama3Dict. (\citeauthor{peritiAutomaticallyGeneratedDefinitions2024}) & 21.98 & - & 0.889 \\
        \midrule
        \midrule
        Monosemic only \Oxford{} &  &  &  \\
        \midrule
        \mTT{} + ft (ours) & 1.33 & 0.345 & 0.839 \\
        FLAN-UL2 (\citeauthor{jhiradEvaluatingLargeLanguage2023})$^\diamondsuit$  & 2.63 & 0.457 & - \\
        \quad + examples & 4.03 & 0.501 & - \\
        \quad + 2 shots & 3.40 & 0.475 & - \\
        \quad + 2 shots + examples & 4.40 & 0.505 & - \\
        \LLama{} (ours) & 0.95 & 0.553 & 0.826 \\
        \quad + 5 shots & 3.91 & 0.537 & 0.851 \\
        \Mistral{} (ours) & 1.18 & 0.557 & 0.837 \\
        \quad + 5 shots & 1.82 & 0.529 & 0.842 \\
        \LLamathree{}{} (ours) & 1.59 & 0.568 & 0.835 \\
        \quad + 5 shots & 3.57 & 0.847 & 0.847 \\
        \quad + ft & 0.57 & 0.511 & 0.806 \\
        LLama3Dict (\citeauthor{peritiAutomaticallyGeneratedDefinitions2024})$^\dagger$ & 0.87 & 0.455 & 0.799 \\
        \bottomrule
        \end{tabular}
    }
    \caption{Comparison of our best models against state-of-the-art approaches on \Oxford{}. Note that for the ``Full \Oxford{}'' section  of the table, we compare against models trained/tested on the entire dataset, which differs substantially from ours. BL, BS and CMT are short for BLEU, BERTScore and COMET, respectively; $^\dagger$ observed results, $^\diamondsuit$ reported by \citet{jhiradEvaluatingLargeLanguage2023}.}
    \label{table:sota}
\end{table}

We compare our results against relevant prior work, focusing on the English language via \Oxford. When it comes to models finetuned on dictionary data, we consider the approaches by \citet{ishiwatari-etal-2019-learning}, who proposed a local-and-global context model based on word embeddings, \citet{reidVCDMLeveragingVariational2020}, who leveraged BERT \cite{devlinBERTPretrainingDeep2019} and combined it with a variational inference framework, \citet{huangDefinitionModellingAppropriate2021}, who propose a specificity-sensitive approach with based on T5 finetunes and re-ranking (RR), and \citet{peritiAutomaticallyGeneratedDefinitions2024}, who finetuned LLMs to generate definitions of words in context. We point out that all of these approaches work with the complete \Oxford{} dataset and rely heavily on example sentences for disambiguation.

For zero-shot and few-shot approaches, we compare our results against the values reported by \citet{jhiradEvaluatingLargeLanguage2023}. To perform an apples-to-apples comparison against \citet{peritiAutomaticallyGeneratedDefinitions2024}, we replicate their setup using our data by finetuning \LLamathree{} in the training portion of the concatenation of our dictionary data. We utilize the same prompt introduced in \S \ref{sec:experimental_setup}, which does not substantially differ from what the authors use, save from the fact that we do not provide example sentences. We follow their training procedure, using 4-bit quantization \cite{dettmers2023qlora} via PEFT with their best hyperparameter configuration.

Table \ref{table:sota} summarizes our comparison against relevant prior work. We see a substantial gap of BLEU scores exists between models tested on the full \Oxford{} dataset, and its monosemic-only subset. These differences are not so significant when looking at COMET or BERTScore. We also note that LLama3Dict, a model trained on dictionary data with example sentences, performs significantly worse in the absence of these examples (BLEU of 21.98 decreased to 0.87). While our similar \LLamathree{} finetune is able to outperform LLama3Dict in this scenario in terms of BERTScore, we see both models perform worse than the base LLMs even when combined with ICL across all metrics. In fact, we observed (Table \ref{table:finetuning_effect_multilingual}) that finetuning on dictionary data, with or without example sentences, decreased performance with respect to the base model in all our studied languages for the monosemic scenario. Finally, our finetuned \mTT{}, although trained only on monosemic data, offers competitive results with the best models in terms of BERTScore.

Overall, we think these results show that the task of deriving the meaning of a word from example sentences differs substantially from the task of eliciting definitions purely by recalling concepts learned during pretraining. For the latter task, which is the focus of this paper, it is still unclear why LLMs finetuned on dictionary data without context sentences do not perform better than base LLMs with ICL. Based on the characterization recently introduced by \citet{panWhatInContextLearning2023}, we think it is possible that both finetuning and ICL lead to improvements in terms of Task Recognition (TR) but not in Task Learning (TL): by being exposed dictionary data, models learn to mimic the \textit{style} in which definitions should be provided, but without actually leading to a better ability to recall definitions. 

Our results can also be explained in light of the discussion by \citet{agarwal2024manyshot}, who point out important limitations of the next-token prediction loss as an indicator of downstream performance. It is possible that the low performance is due to a lack of exposure to lexically-rich data during pretraining, especially for languages other than English, which naive SFT cannot compensate for.

\section{Conclusions}

In this paper, we presented the first multilingual study on definition modeling, introducing monolingual dictionary data for German, Spanish, French and Portuguese. Results show that LLMs can provide generally better results compared to smaller multilingual LMs trained on monosemic dictionary data, but they can be unstable and generate outputs outside the target language. Furthermore, techniques such as ICL and finetuning did not lead to significantly better quality output. Finally, we show that our task can serve as a natural, stable alternative to assess the proficiency of LLMs in a language.

For future work, we would like to incorporate more languages and perform further human evaluation to better understand the relationship between human assessments and automatic metrics. We would also like to tackle the disambiguation problem by collecting usage examples to tackle polysemy. Our supplementary material offers preliminary evidence suggesting that using LLMs could be a viable alternative for this (\S \ref{sec:appendix_polysemy}).

\section*{Acknowledgements}
This paper is supported by the National Institute of Advanced Industrial Science and Technology (AIST) policy-based budget project ``R\&D on Generative AI Foundation Models for the Physical Domain'', and was also partially funded by project JPNP20006, commissioned by the New Energy and Industrial Technology Development Organization (NEDO). The computational resources of the AI Bridging Cloud Infrastructure (ABCI), provided by AIST, were used for most of the experiments in this paper. We are also grateful to the NVIDIA Corporation, which donated one of the GPUs we used. Finally, we would also like to thank our volunteer annotators Daniela Serien, Carolina Dias Alexiou, Mauricio Cordova, Lucie Kunitomo-Jacquin, Antoine Jacquin and Denis Higashi Imoto, for their willingness to help and feedback.
 
\section*{Limitations}

There are a few notable limitations of our work. First, while we tackled five different languages, there is no evidence that our results will generalize to other languages, especially those belonging to different linguistic branches. Furthermore, we recognize that despite their remarkable performance, LLMs may be challenging to access for many researchers, as white-box models require powerful computing resources to run them locally, and black-box models are behind a paywall. Finally, we demonstrated how LLMs can help generate use cases for polysemic words, but have not tackled polysemy.

\section*{Ethics Statement}

Our main objective is to propose a new task to evaluate the abilities of LLMs. One potential use-case is to have a model generate fake definitions that may mislead users that interact with an LLM when deployed.
    
\bibliography{references}

\clearpage

\appendix

\section{Data Sources}
\label{sec:appendix_data}

\subsection{Word Lists}
\label{subsec:appendix_data_wordlists}

\begin{itemize}
    \item For German: \url{https://gist.github.com/MarvinJWendt/2f4f4154b8ae218600eb091a5706b5f4}
    \item For Spanish: \url{https://github.com/keepassxreboot/keepassxc/discussions/9854} (Real Academia Española Corpus de Referencia del Español Actua (RAE, CREA))
    \item For French: \url{https://github.com/kkrypt0nn/wordlists/blob/main/wordlists/languages/french.txt}
    \item For Portuguese: \url{https://natura.di.uminho.pt/download/sources/Dictionaries/wordlists/} 
\end{itemize}

\subsection{Dictionaries}
\label{subsec:appendix_data_dictionaries}

The list below contains links to the Terms of Service for each one of the dictionaries used in our work, as well as relevant data sources from previous work, together with the specific sections where access/distribution rights are stated.

\begin{itemize}
    \item Oxford \cite{gadetsky-etal-2018-conditional}
    \begin{itemize}
        \item \url{https://www.oed.com/information/about-the-oed/legal-notice}
    \end{itemize}
    \item Urban Dictionary \cite{niLearningExplainNonStandard2017a}  
    \begin{itemize}
        \item \url{https://about.urbandictionary.com/tos/} 
    \end{itemize}
    \item Cambridge \cite{ishiwatari-etal-2019-learning} 
    \begin{itemize}
        \item \url{https://www.cambridge.org/legal/copyright} 
    \end{itemize}
    \item Le Petit Robert \cite{reidVCDMLeveragingVariational2020} 
    \begin{itemize}
        \item \url{https://dictionnaire.lerobert.com/mentions-legales}  PROPRIÉTÉ INTELLECTUELLE
    \end{itemize}
    \item Dicio (ours)
    \begin{itemize}
        \item \url{https://www.dicio.com.br/termos-de-uso.html}
    \end{itemize}
    \item DRAE (ours) 
    \begin{itemize}
        \item \url{https://dle.rae.es/contenido/aviso-legal} Point 4
    \end{itemize}
    \item Duden (ours)
    \begin{itemize}
        \item \url{https://www.duden.de/service/agb} Point 11.1
    \end{itemize}
    \item Larousse (ours)
    \begin{itemize}
        \item \url{https://www.larousse.fr/infos/cgu} Section PROPRIÉTÉ INTELLECTUELLE
    \end{itemize}
\end{itemize}

In our paper, we utilize this data in a way that does not necessarily align with the intended use as defined by the providers. As mentioned in \S \ref{sec:data}, in order to make sure we comply with the Terms of Services of our data sources, we will not distribute the definition data we collected, but only the wordlists we compiled. For the same reason, we do not run few-shot experiments with ChatGPT. 

\section{Experimental Setup Details}
\label{sec:appendix_experimental_setup}

\paragraph{Model Training} We train our models with the AdamW optimizer, with an initial learning rate of 5e-5, using a batch size of 16 for \mBART{} or 8 for \mTT{} and for a maximum of 20 epochs. During inference, we sample from the output distribution with a maximum length of 256.

\paragraph{Prompting} Regarding the system prompt, for \LLama{}, we follow the approach by the original paper but replace the original one \footnote{\url{https://github.com/huggingface/transformers/blob/ee2a3400f2a7038a23b83a39c5d0e24f7f699561/src/transformers/models/llama/tokenization_llama.py\#L57}, \url{https://github.com/facebookresearch/llama/blob/main/UPDATES.md}} with the phrase \textit{``You are a helpful assistant. Always answer as helpfully and concisely as possible.''}, to help reduce false refusal rates, i.e., when the model incorrectly refuses to answer a question that it should, for example, due to overly broad instructions to be cautious in how it provides responses. In our case, we observed that this often happened whenever we asked the model to define words that could be offensive in certain contexts. In the case of \Mistral{}, we simply utilize the original system prompt setting. To generate definitions, we sample from the output distribution using \textit{top\_k=40}, \textit{top\_p=0.95}, using a repetition penalty of 1. We generate a maximum number of 256 tokens, which falls well below the length of our gold standard definitions.

\paragraph{Evaluation} To compute the compliance of a language model, i.e. its ability to consistently generate definitions in the required language, we use fasttext-based language classification model \cite{grave-etal-2018-learning}. This is a linear model on top of said embeddings, which recognize 176 languages and was trained on 400 million tokens from Wikipedia, as well as sentences from the Tatoeba website. In terms of performance, this model is superior to previous work, including the readily available ``langid.py'' by \citet{lui-baldwin-2012-langid}, in the TCL, Wikipedia, and EuroGov benchmarks \cite{baldwin-lui-2010-language}, with accuracies of 94.7, 93.0 and 98.7 respectively. 

\section{Detailed Results}

\subsection{Model Training Detailed Results}
\label{subsec:model_training}

In addition to experimenting with mT5, we also trained models based on mBART \cite{liuMultilingualDenoisingPretraining2020,tangMultilingualTranslationExtensible2020}. The original mBART model was pre-trained in denoising full texts in multiple languages. We worked with the ``large'' variation (approx. 340 M parameters) pre-trained on 25 languages (m\textsc{BART-cc25}) \cite{liuMultilingualDenoisingPretraining2020}. Since Portuguese was not included, we also use mBART-50 (m\textsc{BART-50}) \cite{tangMultilingualTranslationExtensible2020}, which was initialized with the former and trained to cover other 25 languages. 

Table \ref{table:results_monolingual} summarizes the results of our monolingual experiments when training models. As we compare the performance of other languages against English, we see that our models can attain similar performance in terms of COMET, with Spanish obtaining the best results. In terms of BLEU scores, we observe an unusually high value for Portuguese, which is due to the limitations of our lemmatization-based data cleaning technique and leads to some data leakage problems. Based on our monolingual finetuning experiments, we utilize \mTT{} as our base model for the multilingual experiments. Table \ref{table:results_multilingual_valid} summarized our results with \mTT{} on the validation split.

\begin{table}[h]  
    \centering
    \scalebox{0.82}{
    \begin{tabular}{c@{\hspace{0.2cm}} l@{\hspace{0.2cm}} c@{\hspace{0.2cm}} c@{\hspace{0.2cm}} c@{\hspace{0.2cm}} c@{\hspace{0.2cm}}}
        \toprule
        \multirow{2}{*}{\bf Dataset} & \multirow{2}{*}{\bf Model} & \multicolumn{2}{c}{\bf Valid} & \multicolumn{2}{c}{\bf Test} \\
        & & \bf BL & \bf CMT & \bf BL & \bf CMT \\
        \midrule
        \multirow{3}{*}{\shortstack{\Duden\\(de)}}
            & \mBARTCC               & 4.06	    & 0.338      & 1.732         & \bf 0.348   \\
            & \mBART                 & \bf 6.47	& 0.338      & \bf 2.477     & 0.347   \\
            & \mTT                   & 4.64	    & \bf 0.343  & 0.939         & 0.346    \\
        \midrule
        \multirow{3}{*}{\shortstack{\Oxford\\(en)}}   
            & \mBARTCC               & 2.08         & 0.340      & 0.00        & \bf 0.345   \\
            & \mBART                 & \bf 3.19	    & 0.340	     & \bf 2.61    & 0.341   \\
            & \mTT                   & 2.71             & \bf 0.341  & 1.33        & \bf 0.345   \\
        \midrule
        \multirow{3}{*}{\shortstack{\DRAE\\(es)}}
            & \mBARTCC               & 7.28        & 0.382      & 7.07       & \bf 0.385   \\
            & \mBART                 & 6.11        & 0.378      & 6.10       & 0.383   \\
            & \mTT                   & \bf 9.19     & \bf 0.379 & \bf 7.70   & 0.384   \\
        \midrule
        \multirow{3}{*}{\shortstack{\Larousse\\(fr)}}
            & \mBARTCC               & 2.59        & \bf0.295   & 2.05       & \bf 0.300   \\
            & \mBART                 & 3.01        & \bf0.295   & \bf 1.73   & 0.298   \\
            & \mTT                   & \bf 3.68    & \bf 0.295  & 0.49       & \bf 0.300    \\
        \midrule
        \multirow{2}{*}{\shortstack{\Dicio\\(pt)}}
            & \mBART                 & 34.02       & 0.316      & \bf 18.58  & 0.351   \\
            & \mTT                   & \bf 44.77    & \bf 0.317  & 11.53  & \bf 0.352   \\
        \bottomrule
    \end{tabular}
    }
    \caption{Results of our monolingual experiments, finetuning multilingual language models on each of our datasets. In the table, CMT is short for COMET and BL is short for BLEU.}
    \label{table:results_monolingual}
\end{table}

\begin{table}[t]
    \centering
    \footnotesize
    \begin{tabular}{clcc}
        \toprule
        \bf Dataset & \bf Training Data & \bf BLEU     & \bf COMET   \\ 
        \midrule
        \multirow{2}{*}{\shortstack{\Duden\\(de)}}
            & de                & \bf 4.640  & 0.343     \\
            & de + en           & 0.497      & \bf 0.345 \\
        \midrule
        \multirow{4}{*}{\shortstack{\Oxford\\(en)}}
            & en                & \bf 2.706  & 0.341     \\
            & en + fr           & 0.040      & \bf 0.347 \\
            & en + de           & 0.775      & \bf 0.347 \\
            & en + fr + es + pt & 0.539      & \bf 0.347 \\
        \midrule
        \multirow{5}{*}{\shortstack{\DRAE\\(es)}}
            & es                & 9.19       & 0.379     \\
            & es + fr           & 8.589      & 0.378     \\
            & es + pt           & \bf 9.519  & \bf 0.380 \\
            & es + pt + fr      & 9.493      & 0.379     \\
            & en + fr + es + pt & 6.365      & \bf 0.380 \\
        \midrule
        \multirow{6}{*}{\shortstack{\Larousse\\(fr)}}
            & fr                & 3.675      & 0.295     \\
            & fr + es           & 2.603      & 0.295     \\
            & fr + pt           & 0.904      & \bf 0.299 \\
            & fr + en           & 0.040      & 0.298     \\
            & fr + es + pt      & \bf 4.353  & 0.294     \\
            & fr + en + es + pt & 1.829      & 0.292     \\
        \midrule
        \multirow{5}{*}{\shortstack{\Dicio\\(pt)}} 
            & pt                & \bf 44.772 & 0.317     \\
            & pt + es           & 41.538     & 0.317     \\
            & pt + fr           & 7.464      & \bf 0.318 \\
            & pt + fr + es      & 43.229     & 0.317     \\
            & en + fr + es + pt & 12.050     & 0.316      \\ 
        \bottomrule
    \end{tabular}
     \caption{Results of our multilingual experiments based on \mTT{} on the validation set, where BL is short for BLEU and CMT is short for COMET. }
    \label{table:results_multilingual_valid}
\end{table}

\subsection{Model Prompting Details}
\label{subsec:model_prompting}

\paragraph{Statistical Significance Details} Table \ref{table:zero_shot_ttest} below shows the details of the p-values for the experiments with LLMs. 

\begin{table}[t]
    \footnotesize
    \begin{tabular}{c@{\hspace{0.25cm}} l@{\hspace{0.25cm}}  c@{\hspace{0.25cm}} c@{\hspace{0.25cm}} c@{\hspace{0.25cm}} c@{\hspace{0.25cm}}}
        \toprule
        \textbf{Dataset} & \textbf{Model} & \bf BL  &  \bf BS  &\bf CMT & \bf Cmp. \\
        \midrule
        \multirow{2}{*}{\shortstack{\Duden\\(de)}} 
            & \LLama & 0.003 & 0.000 & 0.002 & 0.382 \\
            & \Mistral & 0.420 & 0.212 & 0.004 & 0.372 \\
        \midrule
        \multirow{2}{*}{\shortstack{\Oxford\\(en)}}
            & \LLama & 0.000 & 0.000 & 0.037 & - \\
            & \Mistral & 0.008 & 0.002 & 0.002 & - \\
        \midrule
        \multirow{2}{*}{\shortstack{\DRAE\\(es)}}
            & \LLama  & 0.011 & 0.000 & 0.001 & 0.011 \\
            & \Mistral & 0.096 & 0.037 & 0.003 & 0.021 \\
       \midrule
       \multirow{2}{*}{\shortstack{\Larousse\\(fr)}}
            & \LLama   & 0.001 & 0.000 & 0.263 & 0.970 \\
            & \Mistral & 0.003 & 0.000 & 0.040 & 0.775 \\  
       \midrule
       \multirow{2}{*}{\shortstack{\Dicio\\(pt)}}
            & \LLama  & 0.185 & 0.602 & 0.597 & 0.295 \\
            & \Mistral  & 0.027 & 0.012 & 0.035 & 0.000 \\
           \bottomrule
    \end{tabular}
    \caption{P-values of t-test comparing the model's performance in zero and five shot scenarios.}
    \label{table:zero_shot_ttest}
\end{table}

\paragraph{Role of Quantization} Quantization has been shown to be detrimental to performance in certain scenarios, we also evaluate the influence of this technique on our task. For this study, we limit ourselves to the zero-shot scenario and to the \DRAE{}, \Larousse{} and \Dicio{}, which we believe should serve as a reasonable estimate for the overall performance gap due to the effect of quantization. We run each experiments three times with different random seeds, and report average results, which we summarize in Table \ref{table:low_precision}. 

\begin{table}[t]
    \footnotesize
    \centering
    \begin{tabular}{c@{\hspace{0.25cm}} c@{\hspace{0.25cm}}c@{\hspace{0.25cm}} c@{\hspace{0.25cm}} c@{\hspace{0.25cm}}}
        \toprule
        \textbf{Dataset} & \textbf{Model} & \textbf{Type} & \textbf{BL} & \textbf{CMT} \\
        \midrule
        \multirow{4}{*}{\shortstack{\DRAE\\(es)}}
            & \multirow{2}{*}{\LLama}
                & 4bit & 0.271           & 0.504            \\
                & & full & 0.295         & 0.510            \\
            & \multirow{2}{*}{\Mistral}
                 & 4bit & 0.324          & 0.458            \\
                 & & full & 0.333        & 0.469            \\
        \midrule
        \multirow{4}{*}{\shortstack{\Larousse\\(fr)}}
            & \multirow{2}{*}{\LLama}
                & 4bit & 0.326           & 0.473            \\
                & & full & 0.380         & 0.483            \\
            & \multirow{2}{*}{\Mistral}
                & 4bit & 0.327           & 0.441            \\
                & & full & 0.432         & 0.463            \\
        \midrule
        \multirow{4}{*}{\shortstack{\Dicio\\(pt)}}
            & \multirow{2}{*}{\LLama}
                & 4bit & 0.143           & 0.513            \\
                & & full & 0.120         & 0.515            \\
            & \multirow{2}{*}{\Mistral}
                & 4bit & 0.146           & 0.475            \\
                & & full & 0.138         & 0.489   \\
        \bottomrule
    \end{tabular}
    \caption{Results of our study on the impact of low precision (4 bit). In the table, BL and CMT are short for BLEU and COMET, respectively.}
    \label{table:low_precision}
\end{table}

\paragraph{Hardware} Our \mBART{} and \mTT{} finetuning experiments were performed on a large cluster, where we usually relied on a node with 4 NVIDIA V100 GPUs, or an instance with a single NVIDIA A100 GPU. For inference with LLMs and LLM finetuning, we used nodes with 1 NVIDIA H200 GPU. We spent a total of approximately 3,000 USD in our experimental setup, most of which is due to improving the robustness with of our study by repeating LLM experiments with multiple seeds and to running the multilingual LLM benchmarks. 

\subsection{Correlation with Benchmarks}
\label{sec:detail_correlation_benchmarks}

\begin{table}[h]
    \scalebox{0.85}{
    \begin{tabular}{llccc}
        \toprule
        \textbf{Model} & \textbf{Score} & \textbf{BL} & \textbf{BS} & \textbf{CMT} \\ \midrule
        \multirow{3}{*}{\LLama{}} 
         & HellaSwag & 0.49 & 0.95 & 0.74 \\
         & MMLU & 0.55 & 0.97 & 0.72 \\
         & ARC-Challenge & 0.61 & 0.88 & 0.87 \\
        \midrule
        \multirow{3}{*}{\Mistral{}}
         & HellaSwag & 0.13 & 0.95 & 0.73 \\
         & MMLU & 0.04 & 0.91 & 0.63 \\
         & ARC-Challenge & 0.32 & 0.99 & 0.84 \\
        \midrule
        \multirow{3}{*}{\LLamathree{}}
         & HellaSwag & -0.44 & 0.93 & -0.18 \\
         & MMLU & -0.24 & 0.97 & -0.10 \\
         & ARC-Challenge & -0.83 & 0.71 & -0.25 \\
        \bottomrule 
    \end{tabular}
    }
    \caption{Pearson correlation between performance in our task and multilingual benchmarks for LLMs.}
    \label{table:correlation_details}
\end{table}

\subsection{Finetuninig LLMs}

\begin{table}[h]
\centering
\scalebox{0.9}{
\begin{tabular}{c@{\hspace{0.15cm}} l@{\hspace{0.15cm}} c@{\hspace{0.15cm}} c@{\hspace{0.15cm}} c@{\hspace{0.15cm}}}
\toprule
\textbf{Lang.} & \textbf{Model} & \textbf{BL} & \textbf{BS} & \textbf{CMT} \\ \midrule
\multirow{3}{*}{\shortstack{\Duden\\(de)}} 
    & \LLamathree{} (ours) & 3.261 & 0.676 & 0.580 \\
    & \quad + ft (ours) & 0.794 & 0.585 & 0.411 \\
    & \quad + ft (\citeauthor{peritiAutomaticallyGeneratedDefinitions2024}) & 0.147 & 0.574 & 0.360 \\
    \midrule
\multirow{3}{*}{\shortstack{\DRAE\\(es)}}
    & \LLamathree{} (ours) & 1.855 & 0.670 & 0.545 \\
    & \quad + ft (ours) & 0.634 & 0.584 & 0.460 \\
    & \quad + ft (\citeauthor{peritiAutomaticallyGeneratedDefinitions2024}) & 0.186 & 0.570 & 0.379 \\
    \midrule
\multirow{3}{*}{\shortstack{\Oxford\\(en)}}
    & \LLamathree{} (ours) & 1.588 & 0.568 & 0.835 \\
     & \quad + ft (ours) & 0.568 & 0.511 & 0.806 \\
     & \quad + ft (\citeauthor{peritiAutomaticallyGeneratedDefinitions2024}) & 0.867 & 0.455 & 0.799 \\
     \midrule
\multirow{3}{*}{\shortstack{\Larousse\\(fr)}}
    & \LLamathree{} (ours) & 3.443 & 0.676 & 0.543 \\
     & \quad + ft (ours) & 0.809 & 0.608 & 0.496 \\
     & \quad + ft (\citeauthor{peritiAutomaticallyGeneratedDefinitions2024}) & 0.266 & 0.595 & 0.388 \\
     \midrule
\multirow{3}{*}{\shortstack{\Dicio\\(pt)}} &
    \LLamathree{} (ours) & 0.457 & 0.683 & 0.542 \\
     & \quad + ft (ours) & 1.648 & 0.598 & 0.467 \\
     & \quad + ft (\citeauthor{peritiAutomaticallyGeneratedDefinitions2024}) & 0.219 & 0.592 & 0.393 \\
     \bottomrule 
\end{tabular}
}
\caption{Summary of the performance of our \LLamathree{} model finetuned on monosemic multilingual dictionary data on our datasets, compared against LLamaDict \cite{peritiAutomaticallyGeneratedDefinitions2024} \EMT{Highlight best values}.}
\label{table:finetuning_effect_multilingual}
\end{table}

\section{Human Evaluation Details}
\label{sec:human_evaluation_details}

Table \ref{table:likert_score} shows the details of our Likert evaluation scheme, which includes example generations we used as guidance to define our score categories. Participants were verbally instructed about the task and informed that their answers would only be used for the purposes of this study, and they all agreed to participate on their on volition. They are all members of an Engineering department of a University, with ages ranging between 20 and 40 years old.   

    \begin{table*}[t]
    \scriptsize
    \centering
    \begin{tabular}{p{1cm} p{2cm} p{12cm}}
        \toprule
        \textbf{Category} & \textbf{Details} & \textbf{Example} \\
        \midrule
        Very poor & A completely wrong definition, one that contains only the word, or its English translation. & 
        \textbf{Waterspout (\textit{Tromba d'agua})} 
        \newline Definition: Phenomenon defined by the formation of a mass of vapors over rivers, lakes, seas, generally in the shape of a funnel, a cone with the base facing the clouds. 
        \newline Generated: Sure! ``Waterspout'' is a term used to designate a type of flooding caused by a river flood or storm surge, which can flood coastal or plain areas.
        \\
        Poor       & A definition of a related word.  &  
        \textbf{To sharpen (\textit{Nitidificar})}
        \newline Definition: To make something clear, perceptible, sharp; To clear a substance; the concept became clear with your explanation. 
        \newline Generated: To sharpen is a verb that means ``to enhance'' or ``to beautify.'' It is used to describe the action of improving or enriching something, it can be applied to objects, situations or people. Example: ``The sun on the beach sharpened the sea with its rays''.
        \\      
        Acceptable & It gives an idea of the broad sense of the word & \textbf{Supralegal (\textit{Supralegal})}
        \newline Definition: {[}Legal{]} Above the law; that, hierarchically, occupies a superior position in relation to legislation. 
        \newline Generated: Of course! The term ``supralegal'' in Portuguese means ``beyond the law'' or ``outside the law''. It is used to describe acts or situations that are illegal or that are not subject to any type of legal regulation.
        \\
        Good & Gives the correct word definition, but it has some imprecision (e.g. wrong use cases) & 
        \textbf{To oversize (\textit{Superdimensionar})}
        \newline Definition: Grant large dimensions to; give an exaggerated size, an exaggerated scope to something; to exaggerate the achievements of a government. 
        \newline Generated: Of course! The Portuguese ``oversize'' is a term that is not very common, but can be used to describe an action or process that increases or exaggerates a characteristic or property of something. Most of the time, it is used to criticize situations in which someone is exaggerating or increasing something disproportionately. For example, if someone is exaggerating their ability in a certain subject, it could be said that they are ``oversizing'' their ability.
        \\
        Very good  & Correct definition with no imprecision. & \textbf{To sanitize (\textit{Sanitizar})}
        \newline Definition: To comply with hygiene requirements or recommendations; to clean: To sanitize an environment. 
        \newline Generation: Of course! The Portuguese ``sanitize'' refers to eliminating or removing anything that could be considered unclean or dangerous to health. For example, it may refer to disinfecting surfaces or objects, or eliminating bacteria or viruses from an environment. In short, ``sanitize'' means to purify or disinfect.                                          
        \\
        \bottomrule
    \end{tabular}
    \caption{Explanation of our human evaluation system for generated definitions. Examples are taken from \Dicio{} and manually translated to English by the annotator.}
    \label{table:likert_score}
\end{table*}

As mentioned above, in order to further understand the effect of in-context learning in our task, and the effectiveness of COMET as a metric, we perform an additional experiment by using this metric to select the $50$ best and $50$ worst generations, again resulting in a total of $400$ definitions to be evaluated per annotator. We choose the same random seed for all models. We exclude English from this study as this language has been studied before for our task, and because our initial evaluation shows that performance substantially superior.

\begin{table}[h]
    \footnotesize
    \centering
    \begin{tabular}{l c c} 
        \toprule
        \bf Model & \bf Likert (b/w) & \bf COMET (b/w) \\
        \midrule
        \midrule
        \multicolumn{3}{l}{\Duden{} (de)} \\
        \midrule
        $\;$ \LLama         & 3.060 / 1.100 & 0.670 / 0.241 \\
        $\;\;$ + 5 shots    & 1.400 / 1.040 & 0.767 / 0.194  \\
        \cmidrule{2-3}          
        $\;$ \Mistral       & 2.560 / 1.000 & 0.699 / 0.211 \\
        $\;\;$ + 5 shots    & 1.920 / 1.000 & 0.700 / 0.197 \\
        \midrule
        \multicolumn{3}{l}{\DRAE{} (es)} \\
        \midrule
        $\;$ \LLama         & 4.680 / 1.100 & 0.700 / 0.317  \\
        $\;\;$ + 5 shots    & 4.900 / 1.100 & 0.751 / 0.235 \\
        \cmidrule{2-3}  
        $\;$ \Mistral       & 4.600 / 1.080 & 0.727 / 0.258 \\
        $\;\;$ + 5 shots    & 4.440 / 1.140 & 0.710 / 0.236 \\
        \midrule
        \midrule
        \multicolumn{3}{l}{\Larousse{} (fr)}  \\
        \midrule
        $\;$ \LLama         & 4.280	/ 1.122 & 0.697	/ 0.281 \\
        $\;\;$ + 5 shots    & 4.460	/ 1.260 & 0.736 / 0.239  \\
        \cmidrule{2-3}
        $\;$ \Mistral       & 4.061 / 1.000 & 0.705 / 0.233 \\
        $\;\;$ + 5 shots    & 4.040 / 1.000 & 0.689 / 0.228 \\
        \midrule
        \midrule
        \multicolumn{3}{l}{\Dicio{} (pt)} \\
        \midrule
        $\;$ \LLama         & 3.860 / 1.240 & 0.639 / 0.385 \\
        $\;\;$ + 5 shots    & 4.000 / 1.460 & 0.623 / 0.339 \\
        \cmidrule{2-3}
        $\;$ \Mistral       & 3.260 / 1.140 & 0.577 / 0.320 \\
        $\;\;$ + 5 shots    & 1.780 / 1.180 & 0.530 / 0.306 \\
        \bottomrule 
    \end{tabular}
    \caption{Results of our second human evaluation experiment in terms of average Likert scores for the best 50 / worst 50 denoted as (b/w); and compared to COMET.} 
    \label{table:human_eval_2}
\end{table}

Table~\ref{table:human_eval_2} summarizes the results of this second experiment. We first notice that the effects of in context learning are again not consistent across models and languages. We often find that results for ``Worse'' sets contain rare terms, which we assume should also have relative lower frequency in the training data. Given the nature of our task, we think this shows that although few-shot learning may help the model achieve a more dictionary-like style when generating, it does not elicit better task-solving capabilities for those cases. 

We also see observe that Likert scores on the ``Best'' sets are higher than on the ``Worst'' sets, which suggests that COMET is indeed able to capture the semantic quality of the generated definitions. However, from a qualitative point of few, feedback from our annotators suggest that while \LLama{} generations were more precise and correct, \Mistral{} generations were rather ``Very good'', with rich details, or ``Very Poor'', containing wrong information; however, COMET could not detect such differences. These findings are in agreement with previous observations by \citet{jhiradEvaluatingLargeLanguage2023}, and provide additional evidence to show that the performance of COMET when capturing semantic similarity decreases substantially with length of the inputs.

\section{Performance on multilingual benchmarks}
\label{sec:bebchmark_details}

Table \ref{table:bechmmark_performance} summarizes the result we obtained when running our chosen LLMs on the multilingual benchmarks HellaSwag, MMLU and ARC-Challenge. Following \cite{lai-etal-2023-okapi}, experiments on HellaSwag are zero-shot, and we use 25 shots for MMLU and ARC in languages other than English.

\begin{table*}[t]
    \footnotesize
    \centering
    \begin{tabular}{clccc}
        \toprule
        \bf Language & \bf Model & \bf HellaSwag & \bf MMLU & \bf ARC \\
        \midrule
        \multirow{3}{*}{English (en)}
            & \LLama{} & 0.607 & 0.533 & 0.462 \\
            & \Mistral{} & 0.563 & 0.534 & 0.501 \\
            & \LLamathree{} & 0.591 & 0.681 & 0.515 \\
        \midrule
        \multirow{3}{*}{French (fr)}
            & \LLama{} & 0.460 & 0.439 & 0.418 \\
            & \Mistral & 0.426 & 0.413 & 0.387 \\
            & \LLamathree{} & 0.488 & 0.544 & 0.467 \\
        \midrule
        \multirow{3}{*}{German (de)}
            & \LLama{} & 0.431 & 0.431 & 0.408 \\
            & \Mistral & 0.395 & 0.401 & 0.358 \\
            & \LLamathree{} & 0.468 & 0.529 & 0.442 \\
        \midrule
        \multirow{3}{*}{Portuguese (pt)}
            & \LLama{} & 0.437 & 0.434 & 0.427 \\
            & \Mistral & 0.415 & 0.396 & 0.402 \\
            & \LLamathree{} & 0.487 & 0.541 & 0.487 \\
        \midrule
        \multirow{3}{*}{Spanish (es)}
            & \LLama{} & 0.474 & 0.441 & 0.441 \\
            & \Mistral & 0.433 & 0.414 & 0.393 \\ 
            & \LLamathree{} & 0.506 & 0.548 & 0.492 \\ 
        \bottomrule
        \end{tabular}
    \caption{Observed performance of \LLama{} and \Mistral{} on our three selected multilingual benchmarks, separated by language.}
    \label{table:bechmmark_performance}
    \end{table*}

\begin{table*}[t]
    \centering
    \footnotesize
    \begin{tabular}{cccccc}
        \toprule
        \multirow{2}{*}{\textbf{Lang.}} & \multicolumn{3}{c}{\textbf{Existing Benchmark}} & \multicolumn{2}{c}{\textbf{Ours}} \\ 
        \cmidrule(lr){2-4}  \cmidrule(lr){5-6} 
              & MMLU        & ARC       & HellaSwag & Input   & Exp. Output \\
        \midrule
        fr    & 53,431,809  & 1,489,775 & 735,614   & 60,629  & 105,753     \\
        es    & 54,083,040  & 1,460,328 & 707,524   & 62,971  & 64,925      \\
        pt    & 54,044,351  & 1,459,704 & 707,201   & 231,425 & 285,676     \\
        de    & 54,805,912  & 1,459,108 & 735,190   & 139,289 & 108,314     \\ 
        \midrule
        Total & 216,365,112 & 5,868,915 & 2,885,529 & 494,314 & 564,668     \\
        \bottomrule
    \end{tabular}
    \caption{Length of input (and expected output, where available), in number of tokens (via the \LLama{} tokenizer).}
    \label{table:flops}
\end{table*}

\section{Tackling Polysemy}
\label{sec:appendix_polysemy}

Our main results are limited to the set of terms for which we only have one definition. Though our results suggest that LLMs can, to some extent, generate plausible definitions for words, our empirical study offers no insight into the more challenging scenario of polysemy.
Given the success of our studied LLMs in generating definitions for our selected languages, we propose to utilize LLMs to obtain example sentences for polysemic terms in our data. We focus on Portuguese given the annotators' availability.

We run a pilot study with ChatGPT to generate such example sentences. Concretely, we prompt the model as follows: \textit{In the Portuguese Language, the word '$\{$term$\}$' can mean ``$\{$definition$\}$''. Please give me a sentence in that language where this word is used.}, where $\{$term$\}$ and $\{$definition$\}$ are placeholders for variables denoting a given pair of a term and its corresponding definition. 

For our study, we sample a subset of 59 terms from our data, with a total of 156 different meanings. We utilize the official OpenAI API to feed this data into the model. To assess the viability of this approach, we subject these sentences to a thorough human evaluation. Specifically, we ask a native Brazilian Portuguese speaker to evaluate the generated sentences according to the following simplified Likert scale: 1 - Poor: The sentence uses the term with a different meaning; 2 - Acceptable: The sentence uses the term with the specified meaning, but in the wrong context/sounds unnatural; 3 - Good: The sentence uses the term with the specified meaning.

Our results showed that 75.64\% of the examples we obtained were labeled as ``Good'' by our human evaluator, while only 7.69\% and 16.67\% were labeled as ``Poor'' and ``Acceptable'', respectively. 
 
 \begin{table}[t]
    \centering
    \begin{tabular}{c@{\hspace{0.25cm}} c@{\hspace{0.25cm}} c@{\hspace{0.25cm}}} 
        \toprule
        \bf Poor & \bf Acceptable & \bf Good \\
        \midrule
        7.69 & 16.67 & 75.64 \\
        \bottomrule 
    \end{tabular}
    \caption{Results of the human evaluation of the generated example sentences. The numbers indicate the percentage of sentences evaluated in each category}
    \label{table:example_eval}

\end{table}

The results of this evaluation are in Table~\ref{table:example_eval}. We can see that most of the generated examples received the ``Good'' score, with an average of 2.68, which indicates that using this specific LLM could be an alternative to obtain such example sentences for Brazilian Portuguese.

\end{document}